\pdfoutput=1

\documentclass[11pt]{article}

\usepackage[]{acl}
\usepackage[T1]{fontenc}
\usepackage[utf8]{inputenc}

\usepackage{times}
\usepackage{latexsym}
\usepackage{microtype}
\usepackage{inconsolata}

\usepackage{amsmath, amssymb, bm, amsfonts}
\DeclareMathOperator*{\argmax}{arg\,max}
\DeclareMathOperator*{\argmin}{arg\,min}

\usepackage{graphicx}
\usepackage{subcaption}

\usepackage{algorithm}
\usepackage[noend]{algpseudocode}

\usepackage{array}
\usepackage{multirow}
\usepackage{booktabs}
\usepackage{dcolumn}
\newcolumntype{d}[1]{D{/}{/}{#1}}
\usepackage{colortbl}

\usepackage{xcolor}
\definecolor{darkgreen}{rgb}{0.0, 0.45, 0.0}
\definecolor{darkred}{rgb}{0.5, 0.0, 0.0}
\definecolor{darkblue}{rgb}{0.0, 0.0, 0.5}

\usepackage{stfloats}
\usepackage{mdframed}
\usepackage{lipsum}
\usepackage{arydshln}
\usepackage{paralist}
\usepackage{enumitem}
\usepackage{url}
\usepackage{pifont}
\usepackage{hyperref}

\usepackage{tikz}
\usepackage{pgfplots}
\pgfplotsset{compat=1.18}

\newcommand{\greencheck}{\ensuremath{\text{\textcolor{darkblue}{\ding{51}}}}}
\newcommand{\redx}{\ensuremath{\text{\textcolor{red}{\ding{55}}}}}
\newcommand{\textremarkright}[1]{\textcolor{darkblue}{\textbf{#1}}}
\newcommand{\textremarkwrong}[1]{\textcolor{red}{\textbf{#1}}}

\newcommand{\bgcolor}[2]{%
  \hspace{-1.5pt}%
  \tikz[baseline]\node[fill=#1!20, rounded corners=2pt, anchor=base, inner xsep=0pt, inner ysep=0pt]{\ensuremath{#2}};%
  \hspace{-1.5pt}%
}

\DeclareRobustCommand{\bad}{%
  \begingroup\normalfont
    \raisebox{-0.2em}{%
      \includegraphics[height=0.9em]{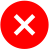}%
    }%
  \endgroup
}
\DeclareRobustCommand{\good}{%
  \begingroup\normalfont
    \raisebox{-0.2em}{%
      \includegraphics[height=0.9em]{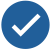}%
    }%
  \endgroup
}

\newcommand{\plotcoords}[4]{%
  (1,#1) (4,#2) (8,#3) (16,#4)
}

\newcommand{\simpleaxis}[6]{%
\begin{axis}[
    xlabel={$n$},
    ylabel={Acc. on $\mathcal{D}_{\text{train}}$},
    xtick={1,4,8,16},
    major tick length=0.1cm,
    tick align=inside,
    width=0.8\linewidth,
    height=0.5\linewidth,
    axis y line*=#4,
    ylabel style={#5},
    axis x line*=bottom,
    axis line style={-},
    ylabel near ticks,#6,
    yticklabel style={/pgf/number format/fixed, /pgf/number format/precision=0},
    legend style={at={(0.5,-0.4)}, anchor=north, legend columns=-1},
    clip=false
]}%

\newcommand{\rightaxis}[1]{%
\begin{axis}[
    ylabel={Acc. on $\mathcal{D}_{\text{eval}}$},
    xtick={1,4,8,16},
    yticklabel pos=right,
    axis x line=none,
    axis y line*=right,
    width=0.8\linewidth,
    height=0.5\linewidth,
    major tick length=0.1cm,
    tick align=inside,
    ylabel style={rotate=-90,xshift=1.5cm},
    ylabel near ticks,
    yticklabel style={/pgf/number format/fixed, /pgf/number format/precision=0},
    clip=false
]}%

\newcommand{\combinedplot}[5]{%
\centering
\begin{tikzpicture}
  \simpleaxis{#1}{Acc. on training set}{south}{left}{align=center}{}
    \addplot[red,mark=x] coordinates {\plotcoords#2};
    \addlegendentry{$\mathcal{D}^{\redx}_{\text{train}}$}
    \addplot[blue,dashed] coordinates {\plotcoords#3};
    \addlegendentry{$\mathcal{D}^{\greencheck}_{\text{train}}$}
    \addplot[black,mark=*] coordinates {\plotcoords#4};
    \addlegendentry{$\mathcal{D}_{\text{train}}$}
    \addlegendimage{black,mark=o}
    \addlegendentry{$\mathcal{D}_{\text{eval}}$}
  \end{axis}

  \rightaxis{Acc. on test set}
    \addplot[black,mark=o] coordinates {\plotcoords#5};
  \end{axis}
\end{tikzpicture}
}

\newcommand{\combinedsubfigure}[6]{%
\begin{subfigure}{\linewidth}
\centering
\combinedplot{#1}{#2}{#3}{#4}{#5}
\caption{#6}
\end{subfigure}
}%

\title{Rectifying Belief Space via Unlearning to Harness LLMs' Reasoning}

\author{Ayana Niwa$^1$  \quad
    Masahiro Kaneko$^1$ \quad
    Kentaro Inui$^{1,2,3}$ \\
    $^1$MBZUAI \quad
    $^2$Tohoku University \quad
    $^3$RIKEN\\
    {\tt \{Ayana.Niwa, Masahiro.Kaneko, Kentaro.Inui\}@mbzuai.ac.ae} \\
    }

\begin{document}
\maketitle
\begin{abstract}
Large language models (LLMs) can exhibit advanced reasoning yet still generate incorrect answers. 
We hypothesize that such errors frequently stem from \textbf{\emph{spurious beliefs}}, which are propositions the model internally considers true but are incorrect. 
To address this, we propose a method to rectify the belief space by suppressing these spurious beliefs while simultaneously enhancing true ones, thus enabling more reliable inferences. 
Our approach first identifies the beliefs that lead to incorrect or correct answers by prompting the model to generate textual explanations, using our \emph{Forward-Backward Beam Search} (FBBS). 
We then apply unlearning to suppress the identified spurious beliefs and enhance the true ones, effectively rectifying the model’s belief space. 
Empirical results on multiple QA datasets and LLMs show that our method corrects previously misanswered questions without harming overall model performance. 
Furthermore, our approach yields improved generalization on unseen data, suggesting that \emph{rectifying a model’s belief space} is a promising direction for mitigating errors and enhancing overall reliability.
\end{abstract}

\section{Introduction} 

Large Language Models (LLMs) trained on massive corpora have demonstrated remarkable reasoning capabilities, even on complex tasks~\cite{NEURIPS2020_1457c0d6,hartmann2023sokmemorizationgeneralpurposelarge,ruis2024proceduralknowledgepretrainingdrives}. 
However,  they still generate logically flawed or factually incorrect answers.
One fundamental question is: why do they generate erroneous outputs, and how can we mitigate such errors?

\begin{figure}[t] 
\centering 
\includegraphics[scale=0.35]{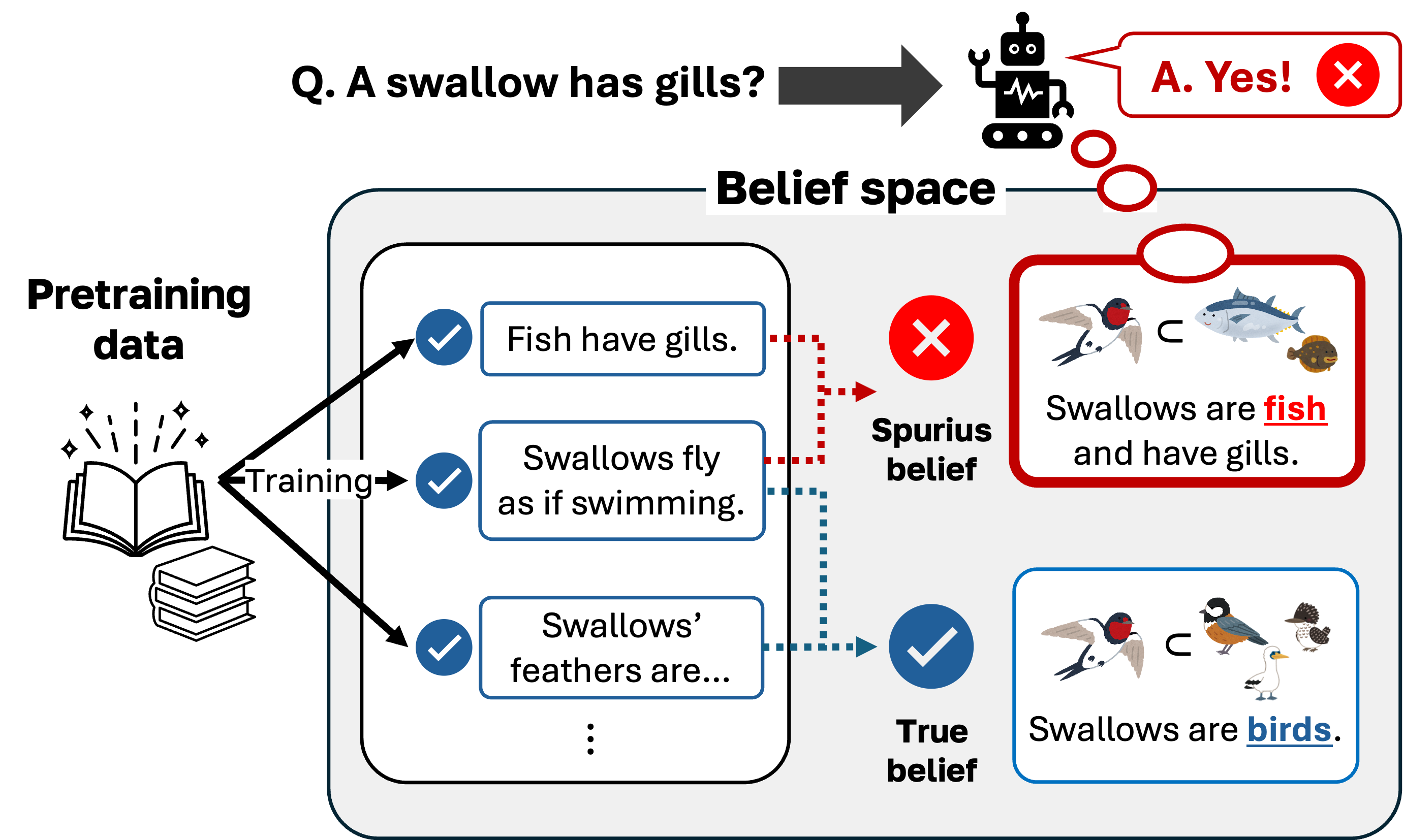} 
\caption{
Example of a QA task: The model combines multiple beliefs from its training data to form new ones. If it references the spurious belief ``\textit{Swallows are fish and ...} \bad'' during its reasoning, it may generate an incorrect answer. This study aims to suppress such spurious beliefs (\bad), thereby allowing the model to draw on the true belief ``\textit{Swallows are birds} \good'' and ultimately avoid erroneous reasoning.
} 
\label{fig:eyecatch} 
\end{figure}

We hypothesize that many of these mistakes arise from \textbf{spurious beliefs} embedded in the model. 
``Belief'' is defined by any proposition the model internally considers true, whereas ``knowledge'' is required to be factually correct~\cite{kassner-etal-2021-beliefbank,richardson-etal-2022-breakpoint,kassner-etal-2023-language}. 
Crucially, LLMs do not merely acquire factual knowledge from training data; rather, they integrate and generalize multiple pieces of knowledge to form new beliefs, resulting in the \textbf{belief space}. 

It is important to emphasize that beliefs are formed regardless of whether they are factually correct.  
Consider a conceptual example in \autoref{fig:eyecatch}: a model might mistakenly combine the true belief ``fish have gills \good'' with ``swallows fly as if swimming \good'' and, could yield a spurious belief such as ``\textit{Swallows are fish and have gills} \bad.’’
If the model references this spurious belief, it may incorrectly answer ``\textit{Yes}’’ to the question ``\textit{Do swallows have gills?}’’~\cite{kassner-etal-2021-beliefbank}. 
This example illustrates how an LLM could form \emph{incorrect implicit beliefs} that are not stated directly in the training corpus.
Our ultimate goal is to rectify the belief space into a more \textbf{trusted space} by suppressing spurious beliefs and enhancing true ones (e.g., \textit{Swallows are birds}’’ \good), preventing incorrect inferences.

In this paper, we propose a framework to rectify the belief space by identifying the beliefs used for the reasoning, and then suppressing references to the spurious beliefs while enhancing references to the true beliefs. 
To identify the beliefs referenced by the model, we instruct it to explain the information required to generate the answer \(y\) from the given input text \(x\).
Specifically, we introduce a Forward-Backward Beam Search (FBBS) that maximizes both the forward likelihood (i.e., the plausibility of the belief given \(x\)) and backward likelihood (i.e., the probability of generating \(y\) from \(x\) and these beliefs) (\autoref{subsec:belief_acquisition}). 
Subsequently, we apply unlearning based on gradient ascent~\cite{yao-etal-2024-machine,Liu2024RethinkingMU} to the identified beliefs to suppress references to spurious beliefs while giving priority to the true ones (\autoref{subsec:unlearning}). 
Through these steps, the model’s belief space is more accurately reorganized, thereby reducing erroneous reasoning.

We demonstrate the effectiveness of our framework on multiple QA tasks (HotpotQA~\cite{yang-etal-2018-hotpotqa}, SciQ~\cite{welbl-etal-2017-crowdsourcing}, and OpenBookQA~\cite{mihaylov-etal-2018-suit}) using three publicly available instruction-tuned LLMs (OLMo~\cite{groeneveld-etal-2024-olmo}, Pythia~\cite{DBLP:conf/icml/BidermanSABOHKP23}, and RedPajama~\cite{weber2024redpajama}). 
Compared to both the vanilla model (before our method) and baseline approaches that either suppress the incorrect answers themselves or knowledge in the training data, our method improves accuracy by up to 6.4 points for OLMo, 5.2 points for Pythia, and 8.0 points for RedPajama.
Moreover, on unseen evaluation data, it achieves gains of up to 9.6 points for OLMo, 7.1 points for Pythia, and 8.4 points for RedPajama, underscoring its strong generalization capability. 
These results notably surpass the vanilla model’s performance, indicating that \textit{rectifying the belief space} can substantially enhance the model’s reasoning. 
Furthermore, suppressing or enhancing beliefs does more than simply target individual beliefs; it effectively reorganizes the entire belief space to reduce errors and improve overall generalizability.

\section{Beliefs in LLMs}
\subsection{Definition of Beliefs}
Following prior work~\cite{kassner-etal-2021-beliefbank,richardson-etal-2022-breakpoint,kassner-etal-2023-language}, we define a \textbf{belief} in an LLM as a proposition that the model \emph{considers} to be true, regardless of whether it is factually correct.
Unlike knowledge, which is generally treated as necessarily factual, beliefs can be erroneous.
We refer to the model's entire collection of such propositions as its \emph{belief space}, denoted by~\(\mathcal{B}\).

Let \(\mathcal{S}\) be the set of all propositions expressed in natural language. We introduce a function \(\Gamma : \mathcal{S} \to \{\mathrm{True}, \mathrm{False}\}\) to determine whether an LLM considers any proposition \(b \in \mathcal{S}\) to be true. 

\begin{align}
\mathcal{B}
&=
\{\,b \in \mathcal{S} \mid \Gamma(b) = \mathrm{True}\},  \\
\Gamma(b)
&=
\begin{cases}
\mathrm{True} & \text{(if the LLM considers \(b\) true)},\\
\mathrm{False} & \text{(otherwise)}.
\end{cases}    
\end{align}

Any belief \(b \in \mathcal{B}\) defined in this way can be categorized into the following two types: 

\paragraph{(1) Explicit Beliefs}
These are propositions that appear directly in the training data and are internalized by the model as-is.
Indeed, numerous studies have shown that LLMs can memorize parts of their training data~\cite{antoniades2024generalization,chen-etal-2024-copybench}, and such memorized content is retained as explicit beliefs within the model.

\paragraph{(2) Implicit Beliefs}
These are propositions that the model internally reconstructs by combining pieces of information or performing analogical reasoning.  
For instance, as shown in \autoref{fig:eyecatch}, the model might derive the belief ``\textit{Swallows are fish and ...} \bad'' by combining information such as ``\textit{Fish have gills} \good’’ and ``\textit{Swallows fly as if swimming} \good.''
Previous work has demonstrated that LLMs are capable of integrating multiple pieces of knowledge for the inference~\cite{treutlein2024connectingdotsllmsinfer}. 
Crucially, even if the original training data is correct, the model may arrive at an spurious belief, leading to incorrect answers.

\subsection{Reasoning Based on Beliefs}
When given an input \(x\) (e.g., a question) and generating an output \(y\) (e.g., an answer), an LLM references some subset \(\mathcal{B}_x \subseteq \mathcal{B}\) of the overall belief space that it deems necessary to answer \(x\). 
From this subset, the model then chooses the most appropriate text \(y\) (i.e., performs inference). 
Formally:

\begin{align}
    y^{*} \;=\; 
    \argmax_{y} 
        P\bigl(y \mid x, \mathcal{B}_x\bigr).
    \label{eq:reasoning_with_beliefs}
\end{align}

We denote by \(\mathcal{B}_{x \rightarrow y}\subseteq \mathcal{B}_x\) the set of beliefs that actually contribute to generating the output \(y\) given the input \(x\). 
In many cases, if \(\mathcal{B}_{x\rightarrow y}\) is factually correct, the model arrives at a correct answer; if \(\mathcal{B}_{x\rightarrow y}\) is spurious, it yields an incorrect answer.
Hence, to reduce erroneous reasoning, we seek to suppress \textbf{spurious beliefs} that lead to mistakes, and rectify the model’s belief space into a more accurate trusted space.

\section{Rectifying the Belief Space}
We propose a two-phase procedure to rectify the belief space \(\mathcal{B}\). 
First, we \emph{identify} which beliefs the model relies on when it generates answers (\autoref{subsec:belief_acquisition} and \ref{subsec:generation}). 
Second, we apply an \emph{unlearning} step to suppress references to spurious beliefs while enhancing references to true ones (\autoref{subsec:unlearning}).

Here, we denote the spurious belief set as \( \mathcal{B}^{\text{Spu}}_{x\rightarrow y_{\text{Inc}}} \) which leads to the incorrect answer \(y_{\text{Inc}}\), and the true belief set as \( \mathcal{B}^{\text{True}}_{x\rightarrow y_{\text{Cor}}} \) which yields the correct answer \(y_{\text{Cor}}\).

\subsection{Identifying Beliefs}
\label{subsec:belief_acquisition}
Consider a given input-output pair \((x, y)\) and the task of identifying the beliefs \(\mathcal{B}_{x\rightarrow y}\) used to derive \(y\) from \(x\). 
Previous research has typically provided candidate beliefs to the model and checked whether the model deems these beliefs true~\cite{kassner-etal-2023-language}. 
However, such an approach does not directly capture \(\mathcal{B}_{x\rightarrow y}\), the set of beliefs specifically used in the inference process from \(x\) to \(y\).

To address this, we adopt an approach based on \textit{explanations}. 
That is, we prompt the model itself, under parameters \(\boldsymbol{\theta}\), to explain which beliefs are necessary to derive \(y\) from \(x\), thereby obtaining the belief set \(\mathcal{B}_{x\rightarrow y}^*\).
Specifically, we adopt a prompt that includes the input \(x\) and the output \(y\), but leaves a blank (represented as \texttt{\_\_\_\_\_}).
By generating the text that fills in this blank, we can acquire the beliefs \(\mathcal{B}_{x \rightarrow y}^*\) that the model itself deems necessary to derive \(x\) to \(y\).
The prompt is as follows\footnote{As mentioned earlier, a belief does not necessarily correspond to an actual fact, but since it is information the model itself considers true, we use the term ``fact'' in the prompt.}:

\begin{mdframed}[
    backgroundcolor=black!5,
    topline=false, bottomline=false, rightline=false, leftline=false,
    innertopmargin=0.5em,
    innerleftmargin=0.5em,
    innerbottommargin=0.5em,
    innerrightmargin=0.5em
]
\begin{quote}
\small
\texttt{\{INPUT\} The concise fact to solve the problem is that \_\_\_\_\_. Therefore, the answer is \{OUTPUT\}.}
\label{prompt}
\end{quote}
\end{mdframed}

We replace \texttt{\{INPUT\}} with \(x\) in the portion preceding the blank to form a prefix prompt \(x^{\text{pre}}\), and replace \texttt{\{OUTPUT\}} with \(y\) in the portion following the blank to form a suffix prompt \(y^{\text{suf}}\).  
To obtain the spurious belief set \(\mathcal{B}_{x\rightarrow y_{\text{Inc}}}^{\text{Spu}}\), we use the prompt by inserting the model’s actual incorrect answer \(y_{\text{Inc}}\) into the \texttt{\{OUTPUT\}} slot.  
Similarly, for the true belief set \(\mathcal{B}_{x\rightarrow y_{\text{Cor}}}^{\text{True}}\), we substitute the correct answer \(y_{\text{Cor}}\) into \texttt{\{OUTPUT\}} slot.

\subsection{Forward-Backward Beam Search (FBBS) for Belief Generation}
\label{subsec:generation}

\begin{figure*}[t]
    \centering
    \small
    \includegraphics[width=\linewidth]{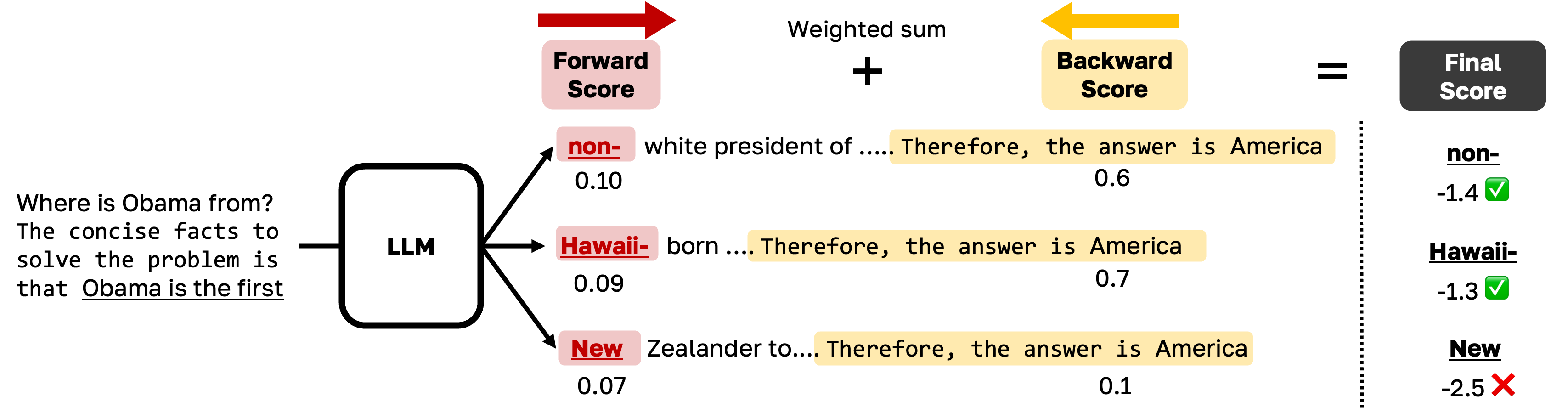}
    \caption{Forward-Backward Beam Search (FBBS) for generating beliefs. 
    Shown here is the process of determining the next token among two candidates, ``non-'' and ``Hawaii-'', (i.e., \(m=2\)) after partially generating the text ``Obama is the first.''}
    \label{fig:beam_search}
\end{figure*}

To generate the beliefs \(b\in\mathcal{B}_{x \rightarrow y}^*\) that fills the blank in the prompt (\autoref{subsec:belief_acquisition}), we must consider both a \bgcolor{red}{\text{forward}} constraint (i.e., plausibility of continuing from \(x^{\text{pre}}\)) and a \bgcolor{yellow}{\text{backward}} constraint (i.e., how likely \(y^{\text{suf}}\) would be generated from \(x^{\text{pre}}\) and a given belief \(b\)).  
Formally, we consider:

\begin{align}
   & \argmax_{b} P\bigl(y^{\text{suf}}, b \mid x^{\text{pre}}; \boldsymbol{\theta}\bigr)\\\nonumber
    \;&=\;
   \argmax_{b}  \bgcolor{red}{\,\underbrace{P\bigl(b \mid x^{\text{pre}}; \boldsymbol{\theta}\bigr)}_{\text{forward}}}
    \;\cdot\;
    \bgcolor{yellow}{\,\underbrace{P\bigl(y^{\text{suf}} \mid x^{\text{pre}}, b; \boldsymbol{\theta}\bigr)}_{\text{backward}}},
\end{align}
\vspace{7pt}

where the first term assesses the plausibility of generating \(b\) from \(x^{\text{pre}}\) (forward), and the second term assesses how well \(y^{\text{suf}}\) is generated from  given \(x^{\text{pre}}\) and \(b\) (backward).  
We achieve this via our proposed Forward-Backward Beam Search (FBBS), an extended version of beam search (see \autoref{fig:beam_search}). 

We consider the case of generating a belief \(b \in \mathcal{B}_{x \rightarrow y}^*\) of length \(T\), which is represented as \((b_1, b_2, \ldots, b_T)\).
For simplicity, here we denote \(x^{\text{pre}}\) as \(x\) and \(y^{\text{suf}}\) as \(y\) in this section.

In standard beam search, the next token \(b_t\) is chosen by maximizing:
\begin{align}
    b_{t} \gets \argmax_{b_t}
    \log P\bigl(b_t \mid x, b_{<t};\boldsymbol{\theta}\bigr),
    \label{eq:vanilla}
\end{align}
where \(b_{<t}\) is the partially generated token sequence \((b_1, \ldots, b_{t-1})\). However, standard beam search does not explicitly account for whether the final output \(y\) will be generated.  
FBBS overcomes this limitation by looking ahead and evaluating the likelihood of ultimately generating \(y\).

Concretely, we repeat the following steps (1)–(4) for \(t = 1, \ldots, T\), thereby identifying sequences \(b\) that lead to \(y\) from \(x\) with high probability. 

\begin{itemize}[leftmargin=10pt]
    \item \textbf{(1) Candidate Selection Based on Token Probability} 
    As in standard beam search, we obtain the top \(n\) token candidates \(\{b_t^{(i)}\}_{i=1}^n\) for step \(t\) by their local conditional probabilities\(\,\, \bgcolor{red}{P_{\mathrm{fwd}}^{(i)}} = P\bigl(b_t \mid x, b_{<t}; \boldsymbol{\theta}\bigr)\).  
    We refer to \(\bgcolor{red}{\log P_{\mathrm{fwd}}^{(i)}}\) as the \textbf{forward} score.

    \item \textbf{(2) Estimating the Probability of Generating \(y^{\text{suf}}\) via Lookahead}  
    For each candidate \(b_t^{(i)}\), we concatenate it with \((x, b_{<t})\) and greedily generate tokens until reaching the end of a sequence.  
    During this process, we measure the probability \(\bgcolor{yellow}{P_{\mathrm{back}}^{(i)}} = P\bigl(y \mid x, b_{<t}, b_t^{(i)};\boldsymbol{\theta}\bigr)\).  
    We refer to \(\bgcolor{yellow}{\log P_{\mathrm{back}}^{(i)}}\) as the \textbf{backward} score.

    \item \textbf{(3) Re-ranking Using a Weighted Score}  
    We combine the forward score \(\bgcolor{red}{\log P_{\mathrm{fwd}}^{(i)}}\) and the backward score \(\bgcolor{yellow}{\log P_{\mathrm{back}}^{(i)}}\) as follows:
  \begin{align}
    S_t^{(i)} 
    &= \lambda\bigl(t,\hat{T}_{t}^{(i)}\bigr) \cdot \underbrace{\bgcolor{red}{\log P_{\mathrm{fwd}}^{(i)}}}_{\text{forward}} \nonumber\\[1mm]
    &\quad + \bigl(1 - \lambda\bigl(t,\hat{T}_{t}^{(i)}\bigr)\bigr) \cdot \underbrace{\bgcolor{yellow}{\log P_{\mathrm{back}}^{(i)}}}_{\text{backward}}\,, \label{eq:score}\\[1mm]
    \lambda\bigl(t,\hat{T}_{t}^{(i)}\bigr)
    &= \frac{1}{1 + \exp\!\Bigl(\alpha \Bigl(\tfrac{2t}{\hat{T}_{t}^{(i)}} - 1\Bigr)\Bigr)}, \label{eq:sigmoid}
\end{align}
    where \(\lambda\bigl(t,\hat{T}_{t}^{(i)}\bigr)\) is a function that dynamically shifts the weight from the forward score to the backward score as generation progresses and \(\hat{T}_{t}^{(i)}\) is the sequence length generated in step (2).  
    Specifically, early in the generation (small \(t\)), we emphasize the next token probability \(\bgcolor{red}{P_{\mathrm{fwd}}}\) to ensure coherent context; later in the generation (large \(t\)), we emphasize the lookahead probability \(\bgcolor{yellow}{P_{\mathrm{back}}}\) to ensure that the final output \(y\) is likely to be generated.  
    The hyperparameter \(\alpha\) controls the smoothness of the sigmoid function.

    \item \textbf{(4) Candidate Update}
    Based on the re-ranked scores \(S_t^{(i)}\), we select the top \(m\,(< n)\) tokens and proceed to generate \(b_{t+1}\).
\end{itemize}

By applying the FBBS method to the input-output pairs \((x, y_{\text{Cor}})\) and \((x, y_{\text{Inc}})\), we can identify the respective beliefs \(\mathcal{B}^{\text{True}*}_{x\rightarrow y_{\text{Cor}}}\) and \(\mathcal{B}^{\text{Spu}*}_{x\rightarrow y_{\text{Inc}}}\).

\subsection{Rectifying the Belief Space via Unlearning}
\label{subsec:unlearning}
Let \(\boldsymbol{\theta}\) denote the parameters of a pretrained model with belief space \(\mathcal{B}\). 
Suppose we aim to suppress the influence of the spurious beliefs \(\mathcal{B}^{\text{Spu}}_{x\rightarrow y_{\text{Inc}}}\) and to enhance the influence of the true beliefs \(\mathcal{B}^{\text{True}}_{x\rightarrow y_{\text{Cor}}}\) within \(\mathcal{B}\).
When we denote by \(\mathcal{B}_r = \mathcal{B} \setminus \mathcal{B}^{\text{Spu}}_{x\rightarrow y_{\text{Inc}}}\), the remaining set of beliefs, the ideal parameters \(\boldsymbol{\theta}_r^*\) that only retain \(\mathcal{B}_r\) are obtained by:
\begin{align}
    \boldsymbol{\theta}^*_{r} = \argmin_{\boldsymbol{\theta}} L(y, \mathcal{B}_r\mid x; \boldsymbol{\theta}),
\end{align}
where \(L(\cdot)\) is a loss function. 
The goal of unlearning in this context is to obtain parameters \(\boldsymbol{\theta}_r^*\) by effectively suppressing the spurious belief set \(\mathcal{B}^{\text{Spu}}_{x\rightarrow y_{\text{Inc}}}\), so that it makes easier to reference the true belief set \(\mathcal{B}^{\text{True}}_{x\rightarrow y_{\text{Cor}}}\).

Concretely, we apply gradient \emph{ascent}~\cite{Liu2024RethinkingMU} to the set of spurious beliefs \(\mathcal{B}^{\text{Spu}}_{x\rightarrow y_{\text{Inc}}}\).  
While standard gradient \emph{descent} updates \(\boldsymbol{\theta}\) to minimize \(L(\boldsymbol{\theta})\), gradient \emph{ascent} updates \(\boldsymbol{\theta}\) in the reverse direction so as to maximize the loss.
Generally, lowering the generation probability of a belief \(\mathcal{B}_{x\rightarrow y}\) makes it more difficult for the model to reference that belief during inference of \(y\), as indicated by \autoref{eq:reasoning_with_beliefs}.
Simultaneously, we explicitly enhance reference to the true beliefs \(\mathcal{B}^{\text{True}}_{x\rightarrow y_{\text{Cor}}}\) (the beliefs that lead to a correct answer). 
Formally:

\begin{align}
\boldsymbol{\theta}^*_{r} =&\argmax_{\boldsymbol{\theta}} \Bigl(\underbrace{\mathbb{E}_{b_i \in \mathcal{B}^{\text{Spu}}_{x\rightarrow y_{\text{Inc}}}} [ L(y_{\text{Inc}}, b_i \mid x; \boldsymbol{\theta})]}_{\text{suppress}} \notag \\
&\quad - \beta \underbrace{\mathbb{E}_{b_i \in \mathcal{B}^{\text{True}}_{x\rightarrow y_{\text{Cor}}}} [ L(y_{\text{Cor}}, b_i \mid x; \boldsymbol{\theta} )]}_{\text{enhance}}\Bigr),
\label{eq:prob_LLM_MU}
\end{align}

where \(\beta\) balances suppressing \(\mathcal{B}^{\text{Spu}}_{x\rightarrow y_{\text{Inc}}}\) and enhancing \(\mathcal{B}^{\text{True}}_{x\rightarrow y_{\text{Cor}}}\).  
By performing this unlearning step, the model is guided toward a rectified belief space that avoids erroneous reasoning.

\section{Experiments}
In this study, we demonstrate that rectifying the belief space can reduce erroneous reasoning while preserving overall model performance.

\subsection{Experimental Settings}
First, the model \(\boldsymbol{\theta}\) is executed on the task using the training data described later. 
Next, we rectify the belief space using our proposed method to obtain \(\boldsymbol{\theta}_r\).
Finally, we use \(\boldsymbol{\theta}_r\)  to perform inference and analyze the results.
During inference, we employ standard beam search to generate the output text \(y\) for a given input \(x\), without using beliefs.

\paragraph{Models}
We experiment with the following three instruction-tuned LLMs:
\begin{enumerate}
    \item \textbf{OLMo (7B)}\footnote{\texttt{allenai/OLMo-7B-Instruct}}~\cite{groeneveld-etal-2024-olmo}
    \item \textbf{Pythia (6.9B)}\footnote{\texttt{allenai/open-instruct-pythia-6.9b-tulu}}~\cite{DBLP:conf/icml/BidermanSABOHKP23}
    \item \textbf{RedPajama (7B)}\footnote{\texttt{togethercomputer/RedPajama-INCITE-7B-Instruct}}~\cite{weber2024redpajama}\quad (abbreviated as RPJ)
\end{enumerate}

\paragraph{Datasets}
We focus on QA tasks that probe the model's belief, using HotpotQA~\cite{yang-etal-2018-hotpotqa} (free-form QA), SciQ~\cite{welbl-etal-2017-crowdsourcing} (multiple-choice QA),
and OpenBookQA~\cite{mihaylov-etal-2018-suit} (multiple-choice QA).\footnote{In our experiments, we only use the \texttt{question} and \texttt{answer} fields of these datasets, and we do not utilize the \texttt{evidence} field, except for the TDA baseline method described in the ``Baseline'' paragraph.}
The LLM must have encountered each training instance from each dataset during its pretraining in order to unlearn them.
Therefore, we first checked the pretraining corpus to verify that both the \texttt{question} and \texttt{answer} fields of each instance were completely included.
 Only those instances meeting this criterion were selected for the training set \(\mathcal{D}_{\text{train}}\). 
The remaining instances are randomly split in equal proportions to create development \(\mathcal{D}_{\text{dev}}\) and evaluation sets \(\mathcal{D}_{\text{eval}}\).  
Hence, for each dataset, the sizes of \(\mathcal{D}_{\text{train}}\), \(\mathcal{D}_{\text{dev}}\), and \(\mathcal{D}_{\text{eval}}\) are: HotpotQA: 70k, 4k, 4k; SciQ: 9k, 2k, 2k; OpenBookQA: 3k, 1k, 1k.

\paragraph{Baselines}
To mitigate erroneous reasoning, we explore several approaches. In addition to assessing the baseline performance (referred to as ``Vanilla'') before any modifications, we compare three distinct methods.
Each method employs the unlearning process described in \autoref{eq:prob_LLM_MU}, utilizing unique suppressing and enhancing sets:
\begin{itemize}
\item \textbf{Answer space rectifying (Answer-SR): } For a given question, we directly suppress the probability of generating an incorrect answer while enhancing that of the correct answer. This is the most straightforward approach to mitigate erroneous reasoning.
\item \textbf{Knowledge space rectifying (Knowledge-SR): } For a given question, we suppress references to irrelevant knowledge and enhance the knowledge that supports the correct answer. The knowledge is identified from the training data. This method aims to prevent the model from incorrectly referencing the knowledge.
\item \textbf{Belief space rectifying (Belief-SR) (Ours): }  For a given question, we suppress references to spurious beliefs while enhancing the true beliefs that support the correct answer.
\end{itemize}

We now provide a more detailed account of Knowledge-SR. 
In our experiment, we define ``knowledge'' as the factually correct information directly contained in the training data, which we assume the model references as the basis for its reasoning.
When the model's reasoning is incorrect, we assume that the training instances it relied on were referenced in error; these instances form our suppressing set in \autoref{eq:prob_LLM_MU}. 
In contrast, our enhancing set represents the knowledge that should have been referenced. We extract it from the \texttt{evidence} field of each dataset, which is part of the model's pretraining corpus.
To identify which pieces of knowledge the model uses, we apply a Training Data Attribution (TDA) method that finds the training instances most influential to the final outputs. 
Among several TDA techniques~\cite{NEURIPS2020_e6385d39,pmlr-v70-koh17a,isonuma-titov-2024-unlearning}, we focus primarily on UnTrac-Inv~\cite{isonuma-titov-2024-unlearning} in this section, as it achieved the best performance in our preliminary experiments. 
Additional experimental details and results for other TDA methods are provided in the Appendix.

\paragraph{Evaluation Metrics}
 We evaluate performance on both the training set \(\mathcal{D}_{\text{train}}\) and the evaluation set \(\mathcal{D}_{\text{eval}}\).
As the evaluation metric, we use accuracy based on the exact match between the prediction and the reference across all datasets.
 Within \(\mathcal{D}_{\text{train}}\), we distinguish:
 \begin{itemize}
\item \(\mathcal{D}^{\redx}_{\text{train}}\), the set of the training instances answered \textit{incorrectly} by the vanilla model \(\boldsymbol{\theta}\). 
\item \(\mathcal{D}^{\greencheck}_{\text{train}}\), the set of the training instances  answered \textit{correctly} by the vanilla model \(\boldsymbol{\theta}\). 
\item \(\mathcal{D}_{\text{train}}\), the entire training set.  
\end{itemize}

\paragraph{Hyperparameters}
We adopt the following settings based on performance on the development set. 
In our Forward-Backward Beam Search (FBBS; \autoref{subsec:generation}), we use \(\alpha = 0.3\) for the sigmoid-based dynamic weighting, a beam width of \(n = 8\), and a candidate size of \(m = 4\).  
For training, we use Adam with a learning rate of \(5 \times 10^{-5}\), a batch size of 8, and \(\beta = 0.5\) in \autoref{eq:prob_LLM_MU}.  
When unlearning, we choose the belief \(b\) with the highest final score (\autoref{eq:score}) as the target. 
We sample the same number of instances for the suppressing and enhancing set in \autoref{eq:prob_LLM_MU}.  
During inference, we apply the default hyperparameters in the Transformers library~\cite{wolf-etal-2020-transformers}.
For additional details, please refer to Appendix.

\begin{table*}[t]
\small
\centering
\begin{tabular}{lrrrrrrrrrrrr}
\toprule
\rowcolor{gray!30}\multicolumn{13}{c}{\textbf{HotpotQA dataset}\rule[-2mm]{0pt}{6mm}}\\
\multirow{2}{*}{Method} & \multicolumn{4}{c}{OLMo} & \multicolumn{4}{c}{Pythia} & \multicolumn{4}{c}{RPJ} \\
\cmidrule(lr){2-5}\cmidrule(lr){6-9}\cmidrule(lr){10-13}
& \multicolumn{1}{c}{\(\mathcal{D}^{\redx}_{\text{train}}\)} & \multicolumn{1}{c}{\(\mathcal{D}^{\greencheck}_{\text{train}}\)} & \multicolumn{1}{c}{\(\mathcal{D}_{\text{train}}\)} & \multicolumn{1}{c}{\(\mathcal{D}_{\text{eval}}\)} & \multicolumn{1}{c}{\(\mathcal{D}^{\redx}_{\text{train}}\)} & \multicolumn{1}{c}{\(\mathcal{D}^{\greencheck}_{\text{train}}\)} & \multicolumn{1}{c}{\(\mathcal{D}_{\text{train}}\)} & \multicolumn{1}{c}{\(\mathcal{D}_{\text{eval}}\)} & \multicolumn{1}{c}{\(\mathcal{D}^{\redx}_{\text{train}}\)} & \multicolumn{1}{c}{\(\mathcal{D}^{\greencheck}_{\text{train}}\)} & \multicolumn{1}{c}{\(\mathcal{D}_{\text{train}}\)} & \multicolumn{1}{c}{\(\mathcal{D}_{\text{eval}}\)}\\
\midrule
Vanilla  & 0.0 & 100.0 & 93.1 & 42.9 & 0.0 & 100.0 & 86.9 & 34.3 & 0.0 & 100.0 & 87.1 & 36.5\\
Answer-SR & \textbf{92.6} & 93.9 & 93.8 & 39.6 & 86.1 & 89.4 & 88.9 & 31.4 & 87.7 & 85.1 & 85.4 & 34.1\\
Knowledge-SR & 81.0 & 89.6 & 89.0 & 42.9 & 83.7 & 85.6 & 85.3 & 33.5 & 86.9 & 84.0 & 84.3 & 35.6 \\\midrule
Belief-SR (Ours) &86.6 & \underline{\textbf{96.1}} & \underline{\textbf{95.4}} & \underline{\textbf{46.2}} & \underline{\textbf{87.7}} & \textbf{91.0} & \textbf{90.5} & \underline{\textbf{38.5}} & \textbf{88.0} & \underline{\textbf{89.4}} & \textbf{89.2} & \underline{\textbf{38.5}} \\
\midrule
\rowcolor{gray!30}\multicolumn{13}{c}{\textbf{SciQ dataset}\rule[-2mm]{0pt}{6mm}}\\
\multirow{2}{*}{Method} & \multicolumn{4}{c}{OLMo} & \multicolumn{4}{c}{Pythia} & \multicolumn{4}{c}{RPJ} \\
\cmidrule(lr){2-5}\cmidrule(lr){6-9}\cmidrule(lr){10-13}
 & \multicolumn{1}{c}{\(\mathcal{D}^{\redx}_{\text{train}}\)} & \multicolumn{1}{c}{\(\mathcal{D}^{\greencheck}_{\text{train}}\)} & \multicolumn{1}{c}{\(\mathcal{D}_{\text{train}}\)} & \multicolumn{1}{c}{\(\mathcal{D}_{\text{eval}}\)} & \multicolumn{1}{c}{\(\mathcal{D}^{\redx}_{\text{train}}\)} & \multicolumn{1}{c}{\(\mathcal{D}^{\greencheck}_{\text{train}}\)} & \multicolumn{1}{c}{\(\mathcal{D}_{\text{train}}\)} & \multicolumn{1}{c}{\(\mathcal{D}_{\text{eval}}\)} & \multicolumn{1}{c}{\(\mathcal{D}^{\redx}_{\text{train}}\)} & \multicolumn{1}{c}{\(\mathcal{D}^{\greencheck}_{\text{train}}\)} & \multicolumn{1}{c}{\(\mathcal{D}_{\text{train}}\)} & \multicolumn{1}{c}{\(\mathcal{D}_{\text{eval}}\)}\\
\midrule
Vanilla & 0.0 & 100.0 & 94.5 & 68.9 & 0.0 & 100.0 & 91.8 & 57.3 & 0.0 & 100.0 & 89.6 & 48.6 \\
Answer-SR & 90.6 &91.1 &91.0 &62.0 &88.5 &92.0 &91.7 &55.0 &89.0 &91.0 &90.7 &44.2\\
Knowledge-SR & 87.1 & 90.2 & 90.0 & 65.0 & 85.4 & 90.1 & 89.7 & 57.0 & 80.5 & 87.1 & 86.4 & 47.8 \\\midrule
Belief-SR (Ours) &\underline{\textbf{92.8}} & \underline{\textbf{95.4}} & \textbf{95.2} & \underline{\textbf{71.4}} & \underline{\textbf{91.7}} & \underline{\textbf{93.4}} & \underline{\textbf{93.2}} & \underline{\textbf{60.2}} & \textbf{89.3} & \textbf{91.4} & \textbf{91.1} & \underline{\textbf{52.6}} \\
\midrule
\rowcolor{gray!30}\multicolumn{13}{c}{\textbf{OpenBookQA dataset}\rule[-2mm]{0pt}{6mm}}\\
\multirow{2}{*}{Method} & \multicolumn{4}{c}{OLMo} & \multicolumn{4}{c}{Pythia} & \multicolumn{4}{c}{RPJ} \\
\cmidrule(lr){2-5}\cmidrule(lr){6-9}\cmidrule(lr){10-13}
 & \multicolumn{1}{c}{\(\mathcal{D}^{\redx}_{\text{train}}\)} & \multicolumn{1}{c}{\(\mathcal{D}^{\greencheck}_{\text{train}}\)} & \multicolumn{1}{c}{\(\mathcal{D}_{\text{train}}\)} & \multicolumn{1}{c}{\(\mathcal{D}_{\text{eval}}\)} & \multicolumn{1}{c}{\(\mathcal{D}^{\redx}_{\text{train}}\)} & \multicolumn{1}{c}{\(\mathcal{D}^{\greencheck}_{\text{train}}\)} & \multicolumn{1}{c}{\(\mathcal{D}_{\text{train}}\)} & \multicolumn{1}{c}{\(\mathcal{D}_{\text{eval}}\)} & \multicolumn{1}{c}{\(\mathcal{D}^{\redx}_{\text{train}}\)} & \multicolumn{1}{c}{\(\mathcal{D}^{\greencheck}_{\text{train}}\)} & \multicolumn{1}{c}{\(\mathcal{D}_{\text{train}}\)} & \multicolumn{1}{c}{\(\mathcal{D}_{\text{eval}}\)}\\
\midrule
Vanilla & 0.0 & 100.0 & 92.0 & 71.7 & 0.0 & 100.0 & 90.6 & 63.5 & 0.0 & 100.0 & 91.2 & 64.5 \\
Answer-SR & 88.3 &90.5 &90.3 &65.8 &83.1 &88.2 &87.7 &59.3 &85.8 &86.4 &86.3 &61.7\\
Knowledge-SR & 87.9 & 90.9 & 90.6 & 69.6 & 83.3 & 92.0 & 91.1 & 63.8 & 80.1 & 89.0 & 88.2 & 64.0 \\\midrule
Belief-SR (Ours) &\underline{\textbf{93.5}} & \underline{\textbf{94.7}} & \underline{\textbf{94.6}} & \underline{\textbf{75.4}} & \underline{\textbf{90.4}} & \textbf{92.0} & \textbf{91.8} & \underline{\textbf{66.0}} & \underline{\textbf{93.3}} & \underline{\textbf{94.5}} & \underline{\textbf{94.3}} & \underline{\textbf{68.2}} \\
\bottomrule
\end{tabular}
\caption{
Accuracy on three QA datasets. 
\textbf{Bold} indicates the highest score in each subset, and 
\underline{underlined} marks scores statistically superior at \(p=0.01\) by bootstrap sampling compared to the second-best approach. 
}
\label{tab:main_results}
\end{table*}

\subsection{Results}

Our main results are shown in \autoref{tab:main_results}.

\paragraph{Belief-Space Rectification Effectively Suppresses Erroneous Reasoning}
Let us begin with the results on the training data.
We observe that across nearly all datasets, models, and baselines, our proposed method consistently improves accuracy on \(\mathcal{D}^{\redx}_{\text{train}}\), i.e., the previously misanswered instances.  
Compared with other rectification methods, it achieves improvements of up to 5.7 points for OLMo, 7.3 points for Pythia, and 13.2 points.
Additionally, accuracy for the previously correct instances \(\mathcal{D}^{\greencheck}_{\text{train}}\) also increases relative to the baselines, leading to overall gains on the entire training set \(\mathcal{D}_{\text{train}}\) (up to 6.4 points for OLMo, 5.2 for Pythia, and 8.0 for RPJ).  
Moreover, the updated model \(\boldsymbol{\theta}_r\) by our proposed method outperforms the vanilla model \(\boldsymbol{\theta}\) by up to 2.6 points for OLMo, 3.6 points for Pythia, and 3.1 points for RPJ.
These results indicate that \textbf{rectifying the belief space can reduce incorrect reasoning without compromising the model’s overall performance.}

\paragraph{Improving Belief Space Also Improves Generalization}
Turning to the results on the evaluation set \(\mathcal{D}_{\text{eval}}\), the model \(\boldsymbol{\theta}_r\) obtained by our proposed method outperforms both the baselines and the original vanilla model \(\boldsymbol{\theta}\).  
Knowledge-SR methods that suppress the knowledge (i. e., training instance) frequently degrade overall performance.
This indicates that they can unintentionally eliminate valid information that is still useful for answering other questions. 
Our proposed approach avoids this pitfall by selectively suppressing only the ``spurious beliefs'' linked to incorrect reasoning, thus preserving necessary knowledge.
When we compare our method to Answer-SR, we observe that  Answer-SR can overfit to those specific instances and perform worse on \(\mathcal{D}_{\text{eval}}\).
In contrast, suppressing beliefs tied to those incorrect answers is more effective.
This aligns with the established insight that jointly incorporating explanations can improve learning efficiency~\cite{hartmann-sonntag-2022-survey}: by jointly unlearning the beliefs (explanations) associated with errors, we achieve better overall outcomes.
In summary, belief-SR also excels in generalization. 
Rather than focusing on individual beliefs in isolation,  this suggests that the model can \textbf{abstractly identify patterns of ``what to forget'' and reorganize the belief space}, thereby reducing errors without losing essential information.
Additionally, we confirmed that performance on out-of-domain generalization was not degraded through cross-evaluations conducted by swapping evaluation datasets. For details, see Appendix~\ref{app:crossevaluation-results}.

\subsection{Analysis}\label{sec:analysis}

\begin{table}[!t]
\centering
\small
\begin{tabular}{lrrr}
\toprule
\rowcolor{gray!30}\multicolumn{4}{c}{\textbf{HotpotQA dataset}\rule[-2mm]{0pt}{6mm}}\\
& OLMo & Pythia & RPJ \\
\midrule
Belief-SR (Ours) & 30.4 & 20.1 & 27.2\\
Knowledge-SR & 100.0 & 100.0 & 100.0\\
\quad + Para (GPT-4) & 71.3 & 60.1 & 68.9\\
\quad + Para (Claude 3) & 65.1 & 58.8 & 74.5\\
\midrule
\rowcolor{gray!30}\multicolumn{4}{c}{\textbf{SciQ dataset}\rule[-2mm]{0pt}{6mm}}\\
& OLMo & Pythia & RPJ \\
\midrule
Belief-SR (Ours) & 27.7 & 23.6 & 30.1\\
Knowledge-SR & 100.0 & 100.0 & 100.0\\
\quad + Para (GPT-4) & 75.4 & 65.1 & 77.2\\
\quad + Para (Claude 3) & 70.4 & 55.6 & 80.1\\
\midrule
\rowcolor{gray!30}\multicolumn{4}{c}{\textbf{OpenBookQA dataset}\rule[-2mm]{0pt}{6mm}}\\
& OLMo & Pythia & RPJ \\
\midrule
Belief-SR (Ours) & 32.3 & 22.4 & 29.6 \\
Knowledge-SR & 100.0 & 100.0 & 100.0\\
\quad + Para (GPT-4) & 80.1 & 59.9 & 79.6\\
\quad + Para (Claude 3) & 73.8 & 59.1 & 77.4 \\
\bottomrule
\end{tabular}
\caption{
\(n\)-gram overlap ratios between model’s pretraining data and spurious beliefs \(\mathcal{B}_{\text{Spu}}\)  (Ours) vs.\ knowledge  in training data identified by Knowledge-SR, including GPT-4 and Claude 3 paraphrasings.
}
\label{tbl:n_gram_match}
\end{table}

\paragraph{Most Beliefs in the Model Are Newly Formed, Not Memorized}

To investigate differences between the pretraining corpus, we measure the \(n\)-gram overlap between the beliefs we identify and the model’s pretraining corpus. 
Specifically, for each dataset, we measure the maximum \(n\)-gram matching to determine what percentage of the identified beliefs appears in the pretraining data. 
For the \(n\)-gram matching, we employed the high-speed engine, Infini-gram~\cite{liu2024infinigram}.
We also test whether the model’s beliefs might simply be paraphrased from the training data. To do this, we take the training instances identified by the TDA method (UnTrac-Inv), paraphrase them using GPT-4~\cite{openai2024gpt4technicalreport} and Claude 3~\cite{Anthropic2023}, then also measure their \(n\)-gram overlap with the pretraining corpus.
\autoref{tbl:n_gram_match} shows that the beliefs generated by our method overlap with the pretraining data at only 20\%--30\% for any dataset and model, whereas the paraphrased UnTrac-Inv samples exhibit overlaps of 60\%--80\% or more.  
This suggests that our method’s beliefs are not merely memorized or paraphrased from the training data, but rather represent newly constructed information.

\begin{figure*}
    \centering
    \includegraphics[width=\linewidth]{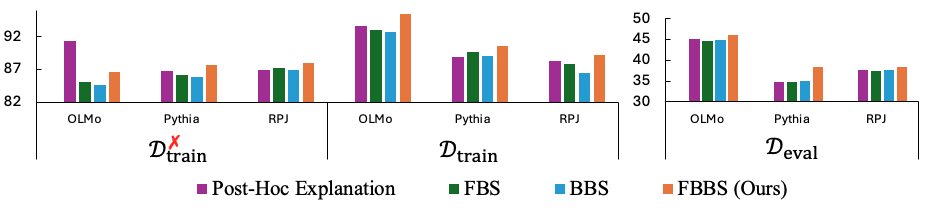}
    \caption{
Comparison of accuracy on HotpotQA when unlearning beliefs generated by various generation methods.
    }
    \label{fig:generation_methods}
\end{figure*}

\begin{figure*}[t]
\centering
\begin{tabular}{cp{10cm}}\toprule
\multirow{2}{*}{\textbf{Question:}} &
     \textit{Which animal has the best camouflage in the Sahara?}\\
     &\textit{(A) a koala bear, (B) a horned viper, (C) Gyrfalcon, (D) a sloth} \\\midrule
\textbf{Correct Answer:}& \textremarkright{(B) \textit{A horned viper}}\\
\textbf{Model Prediction:}& \textremarkwrong{(C) \textit{Gyrfalcon}}\\\midrule
\textbf{Identified Knowledge} & \textit{A desert environment contains very little food} \\
\textbf{Identified Belief \(\mathcal{B}_{\text{Spu}}\)} (Ours)  &  \textit{The gyrfalcon is commonly found in the \textremarkwrong{middle east} and is well-adapted to blending into the sahara's sandy terrain} \bad \\\bottomrule
\end{tabular}
\caption{
An example from the OpenBookQA dataset with OLMo. 
}
\label{tab:output_examples}
\end{figure*}

\paragraph{Forward-Backward Beam Search Delivers Better Overall Accuracy}
We generate beliefs using FBBS described in Section~\ref{subsec:generation}, which jointly optimizes both forward and backward constraints. 
To validate this approach, we compare it against three baseline methods for belief generation:
\begin{description}
    \item[\textbf{Post-Hoc Explanation}] Given input-output pairs \((x, y)\), we prompt an LLM to generate the information needed to derive \(y\) from \(x\).
    \item[\textbf{Forward-only Beam Search (FBS)}] Uses only \(\bgcolor{red}{P_{\mathrm{fwd}}}\), omitting the backward probability.
    \item[\textbf{Backward-only Beam Search (BBS)}] Uses only \(\bgcolor{yellow}{P_{\mathrm{back}}}\), omitting the forward probability.
\end{description}

We performed unlearning the beliefs generated by these baselines and compared their accuracies.

We display the results on HotpotQA in \autoref{fig:generation_methods} (for results on other datasets, see Appendix~\ref{app:all_fbbs_comparison}).
Our FBBS approach achieves the highest overall accuracy on both the training data (\(\mathcal{D}_{\text{train}}\)) and the evaluation data (\(\mathcal{D}_{\text{eval}}\)).
While Post-Hoc Explanation sometimes achieves a higher accuracy on \(\mathcal{D}^{\redx}_{\text{train}}\) (instances previously misanswered), it tends to overfit those instances and degrades performance on the training and evaluation data overall.  
 In contrast, FBBS provides a better balance by generating beliefs that are more broadly correctable.

\paragraph{The Top-1 Belief Sufficiently Represents the Belief Space}

\begin{figure}[t]
\centering

\combinedplot{a}%
{{86.6}{86.4}{85.9}{84.8}}%
{{96.1}{96.9}{95.7}{95.3}}%
{{95.4}{96.2}{95.0}{94.6}}%
{{46.2}{47.3}{46.7}{45.1}}%

\caption{Change in accuracy when increasing the number of beliefs $n$ used for unlearning (HotpotQA).}\label{fig:different_n_hotpotqa}
\end{figure}

We investigated whether increasing the number of identified beliefs (top-$n$, $n=1,4,8,16$) per instance used for rectification would yield additional performance gains.
Figure~\ref{fig:different_n_hotpotqa} presents the results for the HotpotQA dataset (results for other datasets are provided in Appendix~\ref{app:different_n}).
Increasing $n$ does not lead to strictly monotonic improvements; rather, performance typically saturates around $n=1, 4$. 
This is likely due to the additional noise introduced by considering too many beliefs. 
Our findings suggest that the top-1 belief already captures the most critical aggregated information from the belief space, and focusing on a small number of high-impact beliefs is sufficient for significant gains.

\paragraph{Qualitative Evaluation}
We evaluated the plausibility of beliefs generated by the OLMo model. 
For each dataset, we randomly sampled 50 spurious and 50 true beliefs, totaling 300 instances. 
Beliefs were assessed on four criteria, Consistency, Correctness, Conciseness, and Completeness, each rated on a 4-point scale (0–3), where higher scores indicate better evaluation results.
Detailed descriptions of the metric are provided in Appendix~\ref{app:qualitative_metrics}, and results are summarized in Table~\ref{tab:evaluation_results}.
True beliefs ($\mathcal{B}^{\text{True}}$) consistently achieved high scores (mean score of at least 2.3 out of a maximum of 3), indicating that the beliefs generated by our method represent plausible and accurate information relevant to answering the question. 
Conversely, spurious beliefs ($\mathcal{B}^{\text{Spu}}$) received substantially lower Correctness scores, less than half those of true beliefs across all datasets. 
These inaccuracies likely lead to incorrect answers due to errors within these beliefs.

\begin{table}[t]
\centering
\resizebox{\columnwidth}{!}{
\begin{tabular}{c cccc}
\toprule
\textbf{} & \textbf{Consist.} & \textbf{Correct.} & \textbf{Concise.} & \textbf{Complete.} \\
\midrule
\rowcolor{gray!30}\multicolumn{5}{c}{\textbf{HotpotQA dataset}\rule[-2mm]{0pt}{6mm}}\\
$\mathcal{B}^{\text{Spu}}$ & 2.0 & 0.9 & 1.9 & 1.8 \\
$\mathcal{B}^{\text{True}}$ & 2.3 & 2.0 & 2.2 & 2.3 \\
\midrule
\rowcolor{gray!30}\multicolumn{5}{c}{\textbf{SciQA dataset}\rule[-2mm]{0pt}{6mm}}\\
$\mathcal{B}^{\text{Spu}}$& 2.2 & 0.7 & 2.3 & 2.2 \\
$\mathcal{B}^{\text{True}}$ & 2.5 & 2.3 & 2.3 & 2.4 \\
\midrule
\rowcolor{gray!30}\multicolumn{5}{c}{\textbf{OpenBookQA dataset}\rule[-2mm]{0pt}{6mm}}\\
$\mathcal{B}^{\text{Spu}}$ & 2.3 & 1.1 & 2.1 & 2.1 \\
$\mathcal{B}^{\text{True}}$ & 2.6 & 2.5 & 2.2 & 2.4 \\
\bottomrule
\end{tabular}
}
\caption{Manual evaluation results.}
\label{tab:evaluation_results}
\end{table}

\paragraph{Example of the Generated Beliefs}
\autoref{tab:output_examples} presents an example from the OpenBookQA dataset using the OLMo model. 
The model incorrectly predicts that the ``\textit{gyrfalcon},'' which actually inhabits Arctic regions, possesses the best camouflage in the Sahara. 
Knowledge-SR identifies a training instance mentioning merely that ``\textit{a desert environment contains very little food},’’ which fails to explain the model’s wrong inference.  
In contrast, our method explicitly uncovers the spurious belief that ``\textit{The gyrfalcon is commonly found in the Middle East and ...},’’ thus revealing the internal misconception linking falcons to desert environments. 
As such, our approach more accurately pinpoints the faulty reasoning behind the model’s error.

\section{Related Work}
\subsection{Belief Editing in LLMs}

The increased use of LLMs as knowledge bases has driven extensive research into editing the models' beliefs. 
Although existing studies typically use the term ``knowledge editing,'' strictly speaking, these methods modify the LLM's beliefs rather than verified factual knowledge.
Prominent approaches, referred to as knowledge editing~\cite{wang2023knowledge,de-cao-etal-2021-editing,meng2022locating}, directly adjust model parameters to modify these internal beliefs without requiring full retraining. 
Alternative methods update the model’s outputs using external editing networks~\cite{mitchell2022fast} or constrained decoding to suppress outdated beliefs~\cite{sun-etal-2024-outdated}.
However, these studies overlook the fact that the beliefs internally held by the model are not necessarily knowledge, that is, factually correct information.
Our approach significantly differs by explicitly intervening in the model's ``belief space,'' enabling more precise intervention into the model's actual reasoning process.

While several recent studies also address the beliefs of LLMs, their primary goal is often \textit{belief coherence}, ensuring consistency among the beliefs, thus indirectly improving output consistency but not necessarily factual correctness~\cite{kassner-etal-2023-language,wang-etal-2023-chatgpt-defend,jang-etal-2022-becel,kassner-etal-2021-beliefbank}.
In contrast, our research explicitly focuses on \textit{belief factuality}, aiming to improve reasoning accuracy by directly rectifying spurious beliefs through unlearning.
Another novel aspect is that, to achieve this, we enabled identification of beliefs directly linked to specific reasoning.

\subsection{Process Supervision}
Recent research has increasingly recognized the importance of explicitly supervising not only final outputs (\textit{outcome supervision}) but also intermediate reasoning processes (\textit{process supervision}).
For instance, \citet{lightman2024lets} showed that providing  feedback for each intermediate reasoning step notably improves model performance compared to outcome-only supervision. 
Additionally, recent methods have leveraged fine-tuning approaches using explicit reasoning annotations, further enhancing model reasoning capabilities~\cite{ho-etal-2023-large,trung-etal-2024-reft}.
Based on these findings, advanced models such as DeepSeek-R1~\cite{deepseekai2025deepseekr1incentivizingreasoningcapability} explicitly include intermediate reasoning steps in their outputs, indirectly optimizing these steps through reinforcement learning.
However, previous research on unlearning has largely neglected intermediate reasoning processes themselves.
Our study addresses this gap by explicitly investigating the advantages of jointly unlearning beliefs (reasoning processes) alongside final answers. Our results confirm that this combined approach significantly improves the performance.

\section{Conclusion}
In this study, we proposed a method to rectify the belief space by \textbf{selectively suppressing references to spurious beliefs} that lead to erroneous reasoning and \textbf{enhancing references to true beliefs} in the belief space of an LLM. 
Specifically, we identify the beliefs used during inference by prompting the model to explain them, and then we apply unlearning.
Our results demonstrate that our method effectively suppresses spurious beliefs that induce incorrect answers, raising the accuracy on previously misanswered instances without harming overall model performance.
Moreover, we observed improved generalization on unseen data, highlighting the benefits of improving the correctness of the belief space itself. 
These findings show that rectifying the belief space offers a promising approach for both mitigating erroneous reasoning and enhancing the model’s generalization performance.

\section{Limitation}
We introduced the Forward-Backward Beam Search (FBBS) method for generating the belief space of a pretrained model, demonstrating its effectiveness experimentally. However, because FBBS requires lookahead generation at each step, its computational cost is higher than conventional beam search. In practical applications, it would be desirable to develop more efficient search or approximation techniques to reduce this overhead.

Additionally, our experiments were conducted on datasets whose knowledge is contained in the model’s training data, thus restricting the range of dataset-model combinations. Nonetheless, our approach to belief generation can, in principle, be applied to any model for which likelihood scores are available.

\bibliography{acl_latex,anthology}

\appendix
\section{Appendix}
\label{sec:appendix}

\subsection{Further Experimental Details}
\paragraph{Hyperparameter Selection Details}
We performed the hyperparameter search based on development set performance with the following candidate sets: 
\begin{itemize}
  \item \textbf{Forward-Backward Beam Search (FBBS):} The dynamic weighting parameter \(\alpha\) was chosen from \(\{0.3, 0.5, 0.7\}\), with the final value set to 0.3. The beam width and candidate size were fixed at \(n = 8\) and \(m = 4\), respectively.
  \item \textbf{Training:} The learning rate for Adam was selected from \(\{1\times10^{-4}, 5\times10^{-5}, 1\times10^{-5}\}\) and set to \(5 \times 10^{-5}\). The batch size was chosen from \(\{1, 4, 8\}\) and set to 8, while the weight \(\beta\) in \autoref{eq:prob_LLM_MU} was chosen from \(\{0.1, 0.5, 1\}\) and set to 0.5.
\end{itemize}

\paragraph{Details of TDA}
Ideally, TDA methods would examine the entire pretraining corpus to find the most influential training instances. 
However, this is computationally infeasible because the size of the pretraining corpus is massive. 
We therefore restrict the search space to the smaller, dataset-provided \emph{evidence pool}, which still fully contains the relevant knowledge. 
These evidences are guaranteed to be part of the each model’s pretraining data.
We emphasize that this smaller evidence pool is of high quality, containing knowledge that is highly plausible as supporting evidence for the questions. Consequently, restricting TDA to this curated subset does not degrade the baseline TDA methods’ performance.

\subsection{Further Experimental Results}
\subsubsection{Overall Main Results}
We present \autoref{tab:full_results} showing all the results, including those obtained using the several TDA methods (\textbf{Grad-Dot (G-Dot)} \cite{NEURIPS2020_e6385d39}, \textbf{Grad-Cos (G-Cos)} \cite{NEURIPS2020_e6385d39}, \textbf{HIF} \cite{pmlr-v70-koh17a}, \textbf{UnTrac (UT)} \cite{isonuma-titov-2024-unlearning}, and \textbf{UnTrac-Inv (UT-Inv)} \cite{isonuma-titov-2024-unlearning}.
As a result, the effectiveness of our proposed method is still confirmed.

\begin{table*}[t]
\small
\centering
\begin{tabular}{lrrrrrrrrr}
\toprule
\rowcolor{gray!30}\multicolumn{10}{c}{\textbf{HotpotQA dataset}\rule[-2mm]{0pt}{6mm}}\\
& \multicolumn{3}{c}{OLMo} & \multicolumn{3}{c}{Pythia} & \multicolumn{3}{c}{RPJ} \\
\cmidrule(lr){2-4}\cmidrule(lr){5-7}\cmidrule(lr){8-10}
& \(\mathcal{D}^{\redx}_{\text{train}}/\mathcal{D}^{\greencheck}_{\text{train}}\)
& \(\mathcal{D}_{\text{train}}\)
& \(\mathcal{D}_{\text{eval}}\)
& \(\mathcal{D}^{\redx}_{\text{train}}/\mathcal{D}^{\greencheck}_{\text{train}}\)
& \(\mathcal{D}_{\text{train}}\)
& \(\mathcal{D}_{\text{eval}}\)
& \(\mathcal{D}^{\redx}_{\text{train}}/\mathcal{D}^{\greencheck}_{\text{train}}\)
& \(\mathcal{D}_{\text{train}}\)
& \(\mathcal{D}_{\text{eval}}\)\\
\midrule
Vanilla & - & 93.1 & 42.9 & - & 86.9 & 34.3 & - & 87.1 & 36.5\\
G-Dot & 83.9/87.5 & 87.2 & 42.2 & 80.1/85.3 & 84.5 & 31.0 & 83.3/87.2 & 86.6 & 35.6 \\
G-Cos & 83.6/86.7 & 86.4 & 42.7 & 82.6/84.5 & 83.1 & 31.4 & 84.1/87.5 & 87.0 & 36.8 \\
HIF & 84.1/84.0 & 85.8 & 42.2 & 81.4/84.6 & 84.1 & 32.9 & 85.6/86.8 & 86.6 & 37.1 \\
UT & 82.2/88.4 & 87.9 & 42.8 & 83.3/86.1 & 85.7 & 34.1 & 87.4/83.7 & 84.1 & 36.3 \\
UT-Inv & 81.0/89.6 & 89.0 & 42.9 & 83.7/85.6 & 85.3 & 33.5 & 86.9/84.0 & 84.3 & 35.6 \\
Ours &86.6/\textbf{96.1} & \textbf{95.4}&\textbf{46.2}&\underline{\textbf{87.7}}/\textbf{91.0} &\textbf{90.5} &\textbf{38.5} &\textbf{88.0}/\textbf{89.4}&\textbf{89.2} &\textbf{38.5} \\
\midrule
\rowcolor{gray!30}\multicolumn{10}{c}{\textbf{SciQ dataset}\rule[-2mm]{0pt}{6mm}}\\
& \multicolumn{3}{c}{OLMo} & \multicolumn{3}{c}{Pythia} & \multicolumn{3}{c}{RPJ} \\
\cmidrule(lr){2-4}\cmidrule(lr){5-7}\cmidrule(lr){8-10}
& \(\mathcal{D}^{\redx}_{\text{train}}/\mathcal{D}^{\greencheck}_{\text{train}}\)
& \(\mathcal{D}_{\text{train}}\)
& \(\mathcal{D}_{\text{eval}}\)
& \(\mathcal{D}^{\redx}_{\text{train}}/\mathcal{D}^{\greencheck}_{\text{train}}\)
& \(\mathcal{D}_{\text{train}}\)
& \(\mathcal{D}_{\text{eval}}\)
& \(\mathcal{D}^{\redx}_{\text{train}}/\mathcal{D}^{\greencheck}_{\text{train}}\)
& \(\mathcal{D}_{\text{train}}\)
& \(\mathcal{D}_{\text{eval}}\)\\
\midrule
Vanilla & - & 94.5 & 68.9 & - & 91.8 & 57.3 & - & 89.6 & 48.6 \\
G-Dot & 84.9/86.7 & 86.6 & 63.0 & 87.9/91.9 & 91.5 & 57.4 & 85.0/85.4 & 85.3 & 45.9 \\
G-Cos & 88.3/90.1 & 90.0 & 65.9 & 82.7/89.2 & 88.6 & 56.1 & 83.2/85.1 & 84.9 & 45.0 \\
HIF & 89.0/91.0 & 90.8 & 66.0 & 82.0/90.4 & 89.7 & 56.4 & 79.7/85.8 & 85.1 & 45.2 \\
UT & 88.0/89.6 & 89.5 & 65.4 & 87.0/89.6 & 89.3 & 56.7 & 80.7/86.4 & 85.8 & 47.0 \\
UT-Inv & 87.1/90.2 & 90.0 & 65.0 & 85.4/90.1 & 89.7 & 57.0 & 80.5/87.1 & 86.4 & 47.8 \\
Ours &\textbf{92.8}/\textbf{95.4} & \textbf{95.2} & \textbf{71.4} & \textbf{91.7}/\textbf{93.4} & \textbf{93.2} & \textbf{60.2} & \textbf{89.3}/\textbf{91.4} & \textbf{91.1} & \underline{\textbf{52.6}} \\\midrule
\rowcolor{gray!30}\multicolumn{10}{c}{\textbf{OpenBookQA dataset}\rule[-2mm]{0pt}{6mm}}\\
& \multicolumn{3}{c}{OLMo} & \multicolumn{3}{c}{Pythia} & \multicolumn{3}{c}{RPJ} \\
\cmidrule(lr){2-4}\cmidrule(lr){5-7}\cmidrule(lr){8-10}
& \(\mathcal{D}^{\redx}_{\text{train}}/\mathcal{D}^{\greencheck}_{\text{train}}\)
& \(\mathcal{D}_{\text{train}}\)
& \(\mathcal{D}_{\text{eval}}\)
& \(\mathcal{D}^{\redx}_{\text{train}}/\mathcal{D}^{\greencheck}_{\text{train}}\)
& \(\mathcal{D}_{\text{train}}\)
& \(\mathcal{D}_{\text{eval}}\)
& \(\mathcal{D}^{\redx}_{\text{train}}/\mathcal{D}^{\greencheck}_{\text{train}}\)
& \(\mathcal{D}_{\text{train}}\)
& \(\mathcal{D}_{\text{eval}}\)\\
\midrule
Vanilla & - & 92.0 & 71.7 & - & 90.6 & 63.5 & - & 91.2 & 64.5 \\
G-Dot & 85.0/88.9 & 88.5 & 66.7 & 80.9/89.5 & 88.6 & 62.6 & 79.5/88.6 & 87.7 & 62.4 \\
G-Cos & 86.3/90.1 & 89.7 & 67.0 & 85.3/88.3 & 88.0 & 62.6 & 80.2/87.3 & 86.6 & 61.3 \\
HIF & 87.5/90.7 & 90.4 & 67.8 & 84.0/90.2 & 89.6 & 62.9 & 79.6/87.5 & 86.8 & 61.4 \\
UT & 86.4/89.6 & 89.3 & 67.4 & 82.1/91.6 & 90.7 & 63.0 & 78.7/88.1 & 87.2 & 62.1 \\
UT-Inv & 87.9/90.9 & 90.6 & 69.6 & 83.3/92.0 & 91.1 & 63.8 & 80.1/89.0 & 88.2 & 64.0 \\
Ours & \textbf{93.5}/\textbf{94.7} & \textbf{94.6} & \textbf{75.4} & \textbf{90.4}/\textbf{92.0} & \textbf{91.8} & \textbf{66.0} & \textbf{93.3}/\textbf{94.5} & \textbf{94.3} & \underline{\textbf{68.2}} \\
\bottomrule
\end{tabular}
\caption{
Accuracy on three QA datasets for all baselines.
}
\label{tab:full_results}
\end{table*}

\subsubsection{Comparison between Generation Methods}\label{app:all_fbbs_comparison}
\begin{figure*}[t]
    \centering 
\includegraphics[scale=0.5]{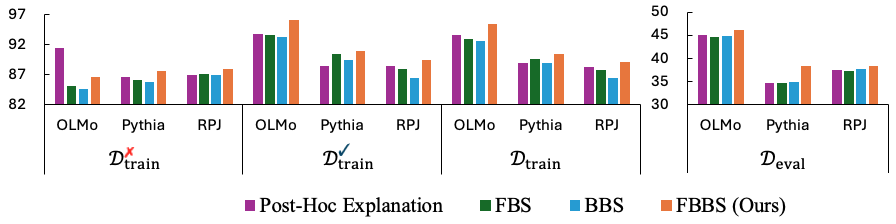} 
\caption{Comparison of belief generation methods on the HotpotQA dataset.} 
\label{app:hotpotqa} 
\end{figure*}

\begin{figure*}[t]
    \centering 
\includegraphics[scale=0.5]{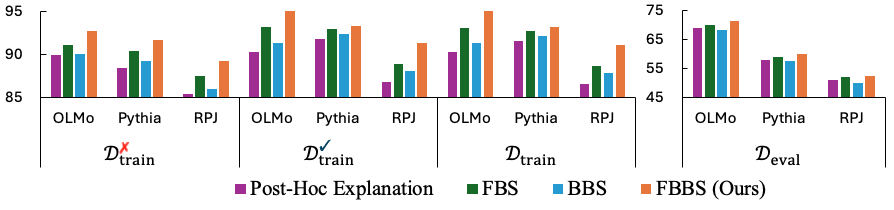} 
\caption{Comparison of belief generation methods on the SciQA dataset. } 
\label{app:sciqa} 
\end{figure*}

\begin{figure*}[t]
    \centering 
\includegraphics[scale=0.5]{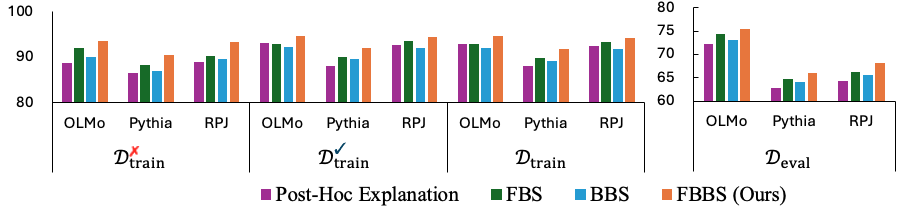} 
\caption{Comparison of belief generation methods on the OpenBookQA dataset.} 
\label{app:openbookqa} 
\end{figure*}

We present the entire results of comparing the multiple belief generation methods introduced in \autoref{sec:analysis}. 
The results for the HotpotQA dataset are shown in \autoref{app:hotpotqa}, those for the SciQA dataset in \autoref{app:sciqa}, and those for the OpenBookQA dataset in \autoref{app:openbookqa}.


\subsubsection{Impact of the Number of Beliefs on Performance}\label{app:different_n}
Figure~\ref{app:fig_different_n} shows how accuracy across all datasets changes when varying the number of beliefs $n$ per example used for rectifying the belief space. In all datasets, performance did not monotonically increase with higher $n$, plateauing at $n= 1$ or $n=4$.
\begin{figure}[t]
\centering
\combinedsubfigure{HotpotQA}%
{{86.6}{86.4}{85.9}{84.8}}%
{{96.1}{96.9}{95.7}{95.3}}%
{{95.4}{96.2}{95.0}{94.6}}%
{{46.2}{47.3}{46.7}{45.1}}%
{HotpotQA}

\combinedsubfigure{SciQA}%
{{92.8}{93.6}{94.0}{93.2}}
{{95.4}{95.8}{94.9}{94.4}}
{{95.2}{95.7}{94.9}{94.3}}
{{71.4}{71.7}{71.0}{70.2}}
{SciQA}

\combinedsubfigure{OpenBookQA}%
{{93.5}{93.4}{92.2}{92.0}}
{{94.7}{95.1}{94.3}{93.6}}
{{94.6}{95.0}{94.1}{93.5}}
{{75.4}{75.7}{74.2}{72.7}}
{OpenBookQA}

\caption{Accuracy with different numbers of the beliefs.}\label{app:fig_different_n}
\end{figure}

\subsubsection{Criteria of Qualitative valuation}\label{app:qualitative_metrics}
The criteria used for the qualitative evaluation of beliefs identified by the proposed method are summarized in Table~\ref{tab:reasoning_criteria}.

\begin{table*}[htbp]
\centering
\begin{tabular}{lp{3cm}p{3cm}p{3cm}p{3cm}}
\toprule
\textbf{Criteria} & \textbf{3} & \textbf{2} & \textbf{1} & \textbf{0} \\
\midrule
Consistency & Logically natural and consistent, with no leaps or contradictions. & Minor leaps or ambiguities, but the overall logic holds. & A clear logical leap or contradiction, making the reasoning insufficient. & Multiple logical leaps or contradictions, rendering the reasoning fundamentally flawed. \\\hline
Correctness & All information is factually accurate. & Some information is ambiguous, but no clear factual errors. & Contains one clear factual error. & Contains multiple factual errors, making content generally unreliable. \\\hline
Conciseness & Includes only necessary information; highly concise. & Slightly redundant, but does not hinder understanding. & Substantial redundancy or irrelevant content, impairing comprehension. & Severely redundant or off-topic, making reasoning difficult to understand. \\\hline
Completeness & All necessary information included. & Some supplementary information missing, but conclusion still reachable. & Lacks important information required to support the conclusion. & Most necessary information missing, making conclusion unsupported. \\
\bottomrule
\end{tabular}
\caption{Manual evaluation criteria for beliefs.}
\label{tab:reasoning_criteria}
\end{table*}

\subsubsection{Cross-evaluation}\label{app:crossevaluation-results}
To evaluate robustness of our proposed method against out-of-domain, we performed a cross-evaluation by swapping evaluation sets among HotpotQA, SciQ, and OpenBookQA. 
As shown in  Table~\ref{tab:crossevaluation-results}, we observed no significant degradation in performance even when the evaluation data was drawn from a different distribution (i.e., another domain).
Naturally, to enhance effectiveness in a new domain, specialized methods such as domain adaptation or transfer learning~\cite{9134370} may be essential.

\begin{table}[htbp]
\centering
\resizebox{\columnwidth}{!}{

\begin{tabular}{lccc}
\toprule
\textbf{Train \textbackslash \, Eval} & \textbf{HotpotQA} & \textbf{SciQA} & \textbf{OpenBookQA} \\
\midrule
\textbf{HotpotQA} & 41.0 & 58.8 & 66.3 \\
\textbf{SciQA} & 38.6 & 61.4 & 67.2 \\
\textbf{OpenBookQA} & 38.4 & 59.1 & 69.8 \\\hline
Vanilla & 37.9 & 58.2 & 66.5 \\
\bottomrule
\end{tabular}
}
\caption{Evaluation results across different combination of training datasets and evaluation datasets.}
\label{tab:crossevaluation-results}
\end{table}

\end{document}